\documentclass[letterpaper]{article}
\usepackage[preprint]{aaai2027}
\usepackage[hyphens]{url}
\usepackage{graphicx}
\usepackage{natbib}
\usepackage{caption}
\usepackage{amsmath,amssymb}
\usepackage{booktabs}
\usepackage{multirow}
\usepackage{colortbl}
\usepackage{array}
\usepackage{algorithm}
\usepackage{algorithmic}
\definecolor{MSABlue}{HTML}{B6D8E4}
\definecolor{MSACoral}{HTML}{FC8D62}
\definecolor{MSAGreen}{HTML}{B1D095}
\definecolor{MSAYellow}{HTML}{EFD985}
\definecolor{MSALightGray}{HTML}{F2F2F2}
\newcommand{\method}{MemOPD}

\title{MemOPD: On-Policy Distillation through Memory State Alignment for Long-Horizon Agents}

\author{
Zhiyuan Liu\textsuperscript{\rm 1}\equalcontrib,
Tinghong Ye\textsuperscript{\rm 2}\equalcontrib,
Chenghao Liu\textsuperscript{\rm 1}\equalcontrib,
Yizhuo Li\textsuperscript{\rm 3},
Songfang Huang\textsuperscript{\rm 1}\corresponding
}
\affiliations{
    \textsuperscript{\rm 1}School of Advanced Manufacturing and Robotics, Peking University\\
    \textsuperscript{\rm 2}Zhejiang University
    \textsuperscript{\rm 3}Harbin Institute of Technology\\
    liuzhiyuan26@stu.pku.edu.cn, yetinghong@zju.edu.cn, chliu@stu.pku.edu.cn,
    yizhuoli815@gmail.com,
    hsf@pku.edu.cn
}
\begin{document}

\maketitle

\begin{abstract}
Long-horizon agents accumulate growing contexts during interaction, impairing performance and stability. Compact memory mitigates this problem by compressing and rewriting the history retained between model invocations. Learning what to retain typically relies on proximal policy optimization (PPO) with final task rewards, but sparse rewards provide little guidance for individual memory updates. This limitation motivates on-policy distillation (OPD), which supplies dense teacher supervision on student rollouts. For such supervision to be valid, the teacher must evaluate each sampled action under the same state in which it was generated. However, the context rewriting performed during memory compression can break this alignment. When sampled responses are retained and re-encoded for later invocations, flattening the interaction into a persistent history may cause the teacher to score the action under a state that the student never visited during rollout. The action therefore remains on-policy by provenance, but not necessarily by state. We therefore propose Memory-Aligned On-Policy Distillation (MemOPD). MemOPD records the inputs and sampled outputs of each model invocation, restores its original token positions and causal visibility, and packs the reconstructed invocations for efficient teacher scoring. The teacher provides full-vocabulary supervision at the sampled action positions, while PPO preserves the final task objective. Experiments verify state alignment across several context updates and show that it improves F1 by 7.0\% over persistent-history teacher scoring in a matched control. Overall, MemOPD-3B improves F1 over PPO by up to 416.2\%, while packing yields up to a $1.63\times$ speedup in actor computation during training. The code for this work is publicly available at: \url{https://github.com/TPssp/MemOPD}.
\end{abstract}

\section{Introduction}

Large language models (LLMs) have driven rapid advances in language agents, enabling them to reason, plan, and interact with external tools~\citep{wang2024survey,yao2023react,nakano2022webgpt}. These agents have demonstrated strong performance in a wide range of long-horizon tasks such as information-seeking workflows, where multiple rounds of reasoning, action execution, and environment feedback are required~\citep{yao2022webshop,deng2023mind2web,gur2024webagent,jin2025search,zheng2025deepresearcher}. During such interactions, each model invocation produces new responses and observations that may influence future decisions. As a result, maintaining the complete interaction history leads to continuously growing contexts, increasing Transformer computation costs and making it more difficult for the model to identify and utilize relevant information~\citep{vaswani2017attention,liu-etal-2024-lost,wu2025longmemeval,an2025effective}.

To address this challenge, existing approaches introduce retrieval mechanisms, external memory systems, or context compression strategies to control the amount of information provided to the model~\citep{lewis_retrieval-augmented_2020,karpukhin-etal-2020-dense,borgeaud_improving_2022,park2023generativeagents,zhongMemoryBankEnhancingLarge2023,packerMemGPTLLMsOperating2024,chhikaraMem0BuildingProductionReady2025a,xuAMEMAgenticMemory2025a,yoon-etal-2024-compact,li2023compressing,cao-etal-2024-retaining,lee2024readagent,jiang2023llmlingua,jiang2024longllmlingua}. Among them, compact memory provides a more direct solution by learning to rewrite the context retained between model invocations, selectively keeping useful information and removing outdated content. For example, MEM1~\citep{zhouMEM1LearningSynergize2025} learns a compact internal state that jointly supports memory consolidation and reasoning, allowing the agent to operate with a bounded context while solving long-horizon tasks.

\begin{figure*}[t]
    \centering
    \includegraphics[width=\textwidth]{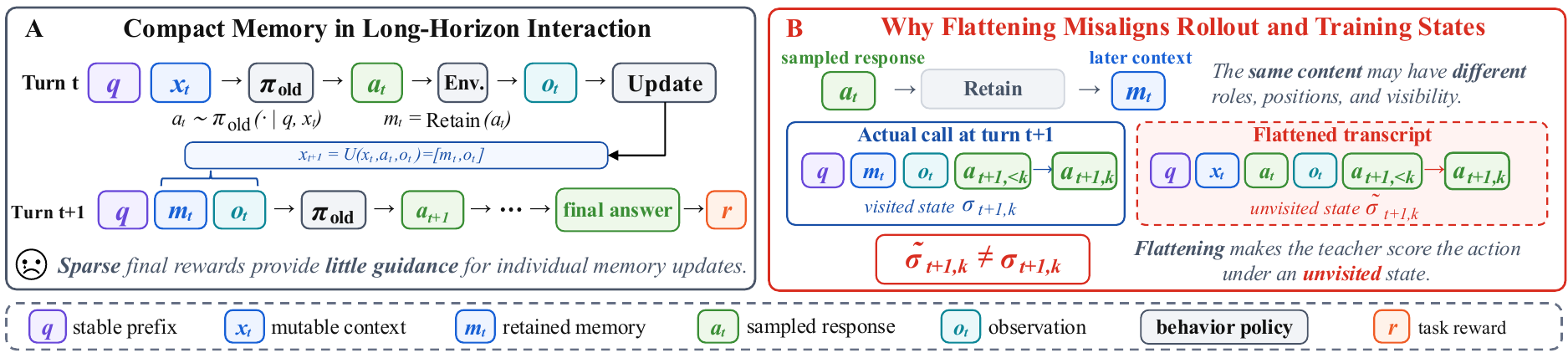}
    \caption{Motivation for \method{}. (A) During compact memory interaction, the behavior policy retains selected content from each response and combines it with the next observation, thereby controlling context growth. PPO learns from the final task reward, which provides limited guidance for individual memory updates. (B) Directly flattening these calls changes the state used to score a sampled action. Content retained as later context can have different roles, positions, and visibility, causing the reconstructed training state $\widetilde{\sigma}_{t+1,k}$ to differ from the rollout state $\sigma_{t+1,k}$. This mismatch motivates memory state alignment.}
    \label{fig:compact-memory-training}
        \vspace{-0.4cm}
\end{figure*}

A common strategy for training such agents is to optimize the entire interaction trajectory with proximal policy optimization (PPO) using the final task reward~\citep{schulman2017proximal,jin2025search,zhouMEM1LearningSynergize2025}. However, the reward is often sparse and delayed, providing limited supervision for individual memory updates.

Teacher distillation provides an alternative source of dense learning signals by transferring the teacher's output distribution to the student model~\citep{hinton2015distilling,kim2016sequencekd}. Recent language model studies further extend distillation to student-generated sequences~\citep{gu2024minillm,ko2024distillm,agarwal2024gkd}. On-policy distillation (OPD) can therefore guide sampled memory decisions while PPO preserves the final task objective.

However, directly applying OPD in this setting introduces an additional requirement: the state used for teacher evaluation must exactly match the state in which the student generated the action. Standard OPD assumes that the autoregressive prefix that produces an action remains unchanged when the action is scored, but context rewriting can break this assumption. During interaction, the agent may generate a response, retain only part of it as memory, and later insert the retained content into a new context. If training simply flattens these interactions into a persistent history, the same response tokens may have different positions, causal visibility, or prediction locations.

Consequently, an action can remain on-policy in terms of its origin, since it was sampled from the student policy, while the reconstructed training state no longer corresponds to any state visited during rollout. The teacher may therefore evaluate an action under an incorrect context, causing the distillation objective to optimize a different conditional distribution from the one that produced the behavior. In our 3B model audit, persistent-history reconstruction produced a p99 log probability error of 1.774, changed the top prediction at 651 sampled action positions, and falsely triggered PPO clipping for 13.29\% of the actions.

This observation motivates Memory-Aligned On-Policy Distillation (MemOPD), which reconstructs the call state that produced each sampled action before applying teacher supervision. Instead of treating the interaction history as a single flattened sequence, \method{} records the exact input and output of every model invocation during rollout, restores the original token positions, causal visibility, and prediction locations, and packs the reconstructed invocations into an efficient training representation. The teacher can then provide dense full vocabulary supervision under the state that produced each action, while PPO continues to optimize the final task objective.

To verify that the reconstruction preserves the original computation, we introduce rollout context equivalence (RCE), which requires the packed representation to produce the same action logits as independently executing each model invocation. RCE therefore tests whether packed training reproduces the conditional computation used during rollout rather than merely forming a valid tensor. Moreover, the reconstructed representation enables efficient batched execution by sharing common computation across invocations without sacrificing state fidelity.

We evaluate \method{} on long-horizon retrieval tasks in which the agent uses compact memory. Across five independently trained seeds, \method{}-3B raises F1 over PPO by up to 416.2\%. In a matched Q2 control, it further improves F1 over persistent-history teacher scoring by 7.0\%. Its packing strategy yields a maximum $1.63\times$ speedup in actor computation during training.

Our contributions are summarized as follows:
\begin{enumerate}
    \item We identify and formulate the state mismatch problem introduced by context rewriting in compact memory agents. We show that student-generated actions alone are insufficient for valid on-policy distillation, and introduce memory state alignment as a necessary condition for applying OPD to long-horizon agents.
    
    \item We propose MemOPD, a training framework that reconstructs each model invocation, separates sampled actions from their later context occurrences, and packs the reconstructed states for efficient teacher-guided optimization.
    
    \item We introduce RCE as a verification criterion for memory-aware training and show that \method{} exactly recovers computation across diverse context-update strategies. Experiments on long-horizon retrieval tasks demonstrate consistent gains in performance and efficiency.

\end{enumerate}

\section{Related Work}
\paragraph{Memory management for long-horizon agents.}
Long-horizon agents combine reasoning with environment actions across web navigation, information seeking, and other multi-turn tasks~\citep{yao2023react,yao2022webshop,nakano2022webgpt,deng2023mind2web,gur2024webagent}. To keep interaction histories bounded, agent systems maintain episodic or hierarchical memories outside the active context~\citep{park2023generativeagents,zhongMemoryBankEnhancingLarge2023,packerMemGPTLLMsOperating2024,chhikaraMem0BuildingProductionReady2025a,xuAMEMAgenticMemory2025a}, while prompt compression and learned reading shorten the context presented to the model~\citep{yoon-etal-2024-compact,lee2024readagent,jiang2023llmlingua,jiang2024longllmlingua}. Other approaches learn context management inside the agent through trajectory folding, explicit memory operations, bounded memory control, or memory specific training signals~\citep{sun2025contextfolding,yu2026agemem,wang2026infmem,li2026contextcurator,zhouMEM1LearningSynergize2025,li2026mempo}. MEM1, for example, learns a consolidated internal state and observes that position changes prevent its packed representation from exactly recovering rollout computation~\citep{zhouMEM1LearningSynergize2025}. These methods mainly study what information an agent should retain, retrieve, or compress. They do not provide the state aligned OPD interface needed to score a sampled decision after its context has been rewritten. \method{} addresses this training condition by recovering the exact state that produced each decision.

\paragraph{Learner states and on-policy distillation.}
Imitation learning has long recognized that learner decisions visit states that differ from a fixed expert trajectory. DAgger therefore queries an expert on states visited by the learner~\citep{ross2011dagger}. This strategy determines where supervision is collected, but it does not ensure that a stored trajectory later reproduces the same conditioning state. Knowledge distillation matches a student to a teacher distribution, while sequence level methods transfer supervision to generated sequences~\citep{hinton2015distilling,gou2021kdsurvey,kim2016sequencekd}. Language model methods further use reverse Kullback--Leibler (KL) objectives or student generated data~\citep{gu2024minillm,ko2024distillm}. Generalized knowledge distillation formulates OPD by querying a teacher on student generated sequences~\citep{agarwal2024gkd}. Standard OPD assumes that the autoregressive prefix used to generate an action remains available when the action is scored. Context rewriting breaks this assumption because a response can later reappear as context with different positions or visibility. Consequently, applying standard OPD directly does not guarantee valid teacher supervision. \method{} instead reconstructs the conditioning state and preserves the sampled action domain; its reverse KL term is unchanged.

\section{Method}
\subsection{Overview}
An agent with compact memory repeatedly invokes a trainable student policy, executes an external action, and rewrites the context for the next invocation. If training flattens these invocations into one persistent sequence, action likelihoods and teacher targets can be evaluated under inputs that the student never encountered. The resulting objective no longer represents the policy trajectory that produced the data, even though the batch remains syntactically valid.

\begin{figure*}[t]
    \centering
    \includegraphics[width=\textwidth]{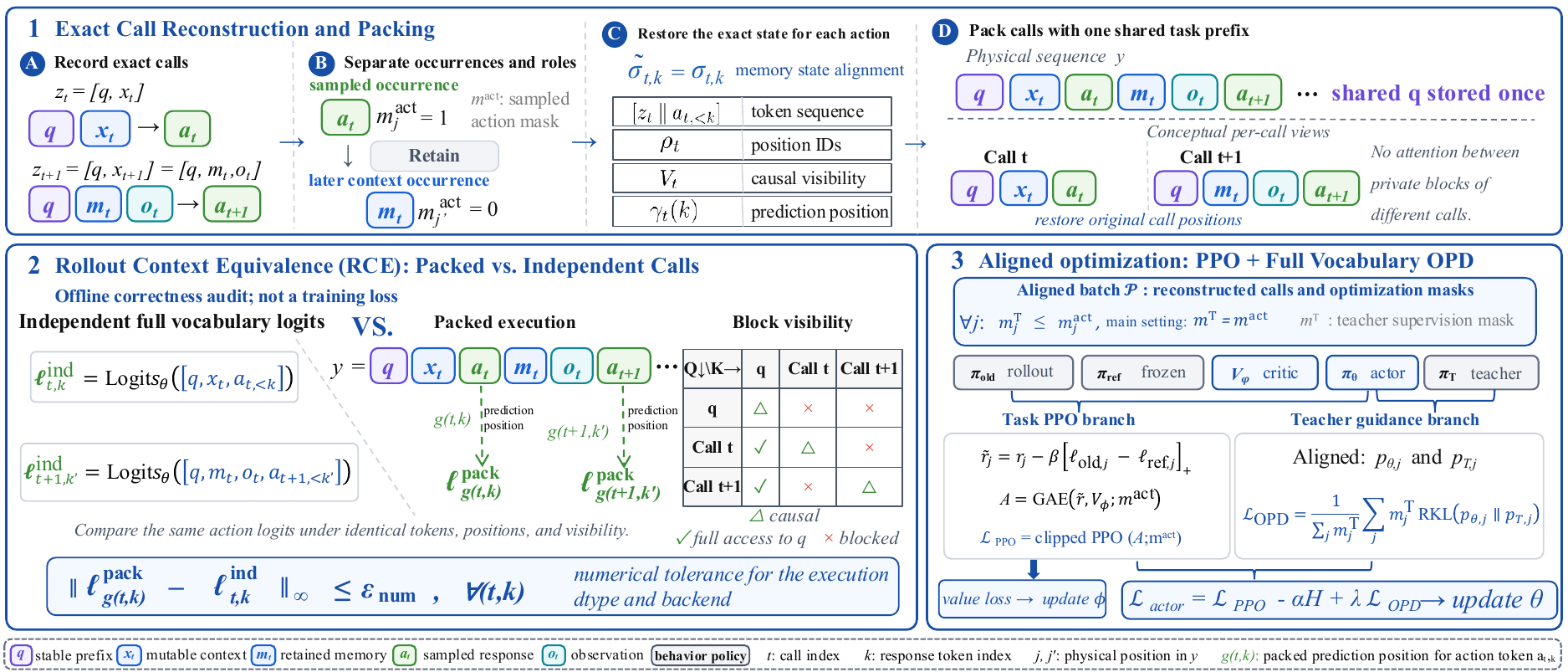}
    \caption{Overview of \method{}. Here, a \emph{call} is a model invocation. (1) MemOPD reconstructs each call, separates sampled actions from later context occurrences, and packs the aligned computations under a shared task prefix. (2) RCE compares packed and independent full vocabulary logits within numerical tolerance. (3) The aligned batch supports PPO over the sampled action domain and full vocabulary teacher guidance over the teacher supervision mask.}
    \vspace{-0.4cm}
    \label{fig:memopd-overview}
\end{figure*}

\method{} prevents this mismatch by recording each student invocation, reconstructing its memory state, and packing the independent computations. At each update, a frozen snapshot $\pi_{\mathrm{old}}$ of the trainable student $\pi_\theta$ generates the rollout, whereas the frozen teacher $\pi_T$ only scores recorded actions after reconstruction. OPD then supplies local guidance on these actions, while PPO retains the task objective. The reference policy $\pi_{\mathrm{ref}}$ and critic $V_\phi$ use the same reconstruction, so every objective uses consistent state and action semantics. Figure~\ref{fig:memopd-overview} illustrates the pipeline.

\subsection{Compact Memory Rollouts}
To identify the state that training must reproduce, we first formalize how compact memory generates each sampled action.
We write the fixed task and system prefix as $q$; it specifies the task and the interaction protocol. Before invocation $t$, the mutable context $x_t$ contains the memory and latest observation available to the student. Conditioned on $[q,x_t]$, the behavior policy samples a response $a_t$ that updates memory and selects a search or final-answer action. The environment executes this action and returns an observation $o_t$, a task reward $r_t$, and a terminal indicator $d_t$. The context update rule $U$ then constructs the input for the next invocation:
\begin{equation}
\begin{aligned}
    a_t &\sim \pi_{\mathrm{old}}(\,\cdot\mid q,x_t),
    & (o_t,r_t,d_t)&=\mathcal E(a_t),\\
    x_{t+1}&=U(x_t,a_t,o_t).&&
\end{aligned}
    \label{eq:transition}
\end{equation}
The student policy, the environment, and $U$ jointly determine the states visited during rollout. The teacher does not generate a separate trajectory; it is queried only after these student states have been reconstructed.

Our end-to-end study instantiates $U$ with the compact-memory protocol of MEM1~\citep{zhouMEM1LearningSynergize2025}. The stable prefix contains the task and interaction instructions. Each response contains reasoning, retained memory, and either a search command or a final answer. After a search, the returned evidence becomes the new observation. The runtime keeps the response content required for the next invocation, appends the new evidence, and removes the earlier mutable context. A three-invocation interaction therefore has the form
\begin{equation}
    [q]\rightarrow a_0,\qquad
    [q,m_0,o_0]\rightarrow a_1,\qquad
    [q,m_1,o_1]\rightarrow a_2,
    \label{eq:mem1-example}
\end{equation}
where $m_t$ is the content retained from response $a_t$. The third invocation receives $[q,m_1,o_1]$, not an accumulated history such as $[q,a_0,o_0,a_1,o_1]$. This distinction is central to training because a sampled response and its later context copy can have different token positions and visibility.

Training first uses supervised fine tuning (SFT) to teach the required reasoning, memory, search, and answer format. The frozen behavior policy $\pi_{\mathrm{old}}$ then samples complete student interactions, and PPO optimizes the student using the final task reward. The reference policy limits drift, while the critic and generalized advantage estimation (GAE) provide token-level learning signals. Before any objective scores an action, \method{} reconstructs the invocation that produced it. The frozen teacher then adds dense supervision without generating the trajectory or replacing the task reward.

\subsection{Memory State Alignment}
\paragraph{Rollout and training states.}
Valid training requires every sampled action to be scored under the invocation that produced it. For invocation $t$, let $z_t=[q,x_t]$ be the exact tokenized input. We define its realized memory state as
\begin{equation}
    \sigma_t=(z_t,\rho_t,V_t,\gamma_t),
    \label{eq:call-state}
\end{equation}
where $\rho_t$ gives token positions, $V_t$ specifies causal visibility, and $\gamma_t(k)$ identifies the position whose output distribution predicts action token $a_{t,k}$. Thus, a memory state is the complete model input presented at an invocation, including its token IDs and computation structure. It is neither the retained text alone nor a Transformer hidden state, and identical decoded text can represent different memory states.

The state used to predict action token $a_{t,k}$ also includes the earlier tokens $a_{t,<k}$ from the same response. Let $\sigma_{t,k}=(z_t,a_{t,<k},\rho_t,V_t,\gamma_t(k))$ denote this rollout state and $\widetilde\sigma_{t,k}$ the state reconstructed during training. Memory state alignment requires $\widetilde\sigma_{t,k}=\sigma_{t,k}$ for every sampled action token. When this equality fails, behavior likelihoods, PPO ratios, and teacher targets no longer describe the sampled decision, so action provenance alone is insufficient.

\paragraph{Reconstruction and packing.}
To enforce this equality, \method{} records $z_t$ and the exact sampled token IDs of $a_t$ during rollout rather than decoding and tokenizing the response again. Keeping these IDs preserves the sampled action and its token boundaries throughout reconstruction.

The reconstruction compiler places the recorded invocations in one physical sequence,
\begin{equation}
    y=[q,\,x_0,a_0,\,x_1,a_1,\ldots,x_{T-1},a_{T-1}],
    \label{eq:packed}
\end{equation}
while treating physical adjacency and causal visibility separately. The stable prefix $q$ has one physical occurrence, keeps positions $0,\ldots,|q|-1$, and is visible to every invocation block. This sharing is valid because its token IDs and positions are identical in all independent executions. A token in $x_t$ sees only $q$ and preceding tokens in the context of invocation $t$. A token in $a_t$ additionally sees preceding tokens of $a_t$, while tokens assigned to other invocations are blocked. Finally, positions restart from the indices used by the corresponding independent invocation instead of increasing across the packed sequence.

When a response is retained for a later invocation, the compiler represents it twice because its two occurrences serve different purposes. The first is the exact sampled action in the policy trajectory. The second is context for a later invocation and therefore receives the positions and visibility of that invocation. If a sequence limit would remove a conditioning token or sampled action, the compiler rejects the example rather than silently changing the training state.

A correct reconstruction should reproduce, for every sampled action token, the same full vocabulary logits as independent execution. We call this property rollout context equivalence (RCE). Let $g(t,k)$ denote the position in the packed sequence whose output distribution predicts action token $a_{t,k}$. Let $\boldsymbol{\ell}^{\mathrm{pack}}_{g(t,k)}$ and $\boldsymbol{\ell}^{\mathrm{ind}}_{t,k}$ be the corresponding full vocabulary logits from packed and independent execution. RCE requires
\begin{equation}
    \left\|
    \boldsymbol{\ell}^{\mathrm{pack}}_{g(t,k)}
    -\boldsymbol{\ell}^{\mathrm{ind}}_{t,k}
    \right\|_\infty
    \leq \epsilon_{\mathrm{num}}
    \quad \text{for every }(t,k),
    \label{eq:rce}
\end{equation}
where $\epsilon_{\mathrm{num}}$ is the numerical tolerance for the execution dtype and backend. We test this condition directly against independent invocation logits for the same recorded actions.

\paragraph{Sampled action domain.}
Correct conditioning does not determine which physical tokens are policy decisions. Retained response copies and environment observations condition future actions, but they were not sampled at those physical positions. We mark these decisions as follows:
\begin{equation}
    m^{\mathrm{act}}_j=
        \mathbf{1}[\,y_j\text{ lies in the sampled occurrence of a recorded }a_t\,].
    \label{eq:action-mask}
\end{equation}
The selected positions form the sampled action domain. Although a later context occurrence may repeat the same token IDs, it occupies different physical positions and receives $m^{\mathrm{act}}=0$. Policy loss, entropy, reference regularization, value learning, and GAE~\citep{Schulman2016GAE} use this domain. GAE advances across ordered sampled actions and skips positions used only for context. Teacher supervision has an indicator $m^{\mathrm{T}}$ satisfying $m^{\mathrm{T}}_j\leq m^{\mathrm{act}}_j$. The teacher mask may select a subset of the action domain; in the main experiment, it covers every sampled response token, so $m^{\mathrm{T}}=m^{\mathrm{act}}$.

\begin{table*}[t]
\centering
\scriptsize
\setlength{\tabcolsep}{2.1pt}
\renewcommand{\arraystretch}{1.04}
\resizebox{\textwidth}{!}{%
\begin{tabular}{lcccccccccccc}
\toprule
\multirow{2}{*}{Model} & \multicolumn{4}{c}{Q2} & \multicolumn{4}{c}{Q8} & \multicolumn{4}{c}{Q16} \\
\cmidrule(lr){2-5}\cmidrule(lr){6-9}\cmidrule(lr){10-13}
& EM $\uparrow$ & F1 $\uparrow$ & Peak $\downarrow$ & Time $\downarrow$
& EM $\uparrow$ & F1 $\uparrow$ & Peak $\downarrow$ & Time $\downarrow$
& EM $\uparrow$ & F1 $\uparrow$ & Peak $\downarrow$ & Time $\downarrow$ \\
\midrule
Qwen2.5-14B-Inst & 0.734 & 0.904 & $15.45\!\pm\!0.17$ & $5.55\!\pm\!0.17$ & 1.552 & 1.872 & $44.52\!\pm\!0.33$ & $16.11\!\pm\!0.25$ & 0.567 & 0.704 & $38.18\!\pm\!0.72$ & $29.63\!\pm\!0.62$ \\
Qwen2.5-7B-Inst & 0.267 & 0.365 & $19.64\!\pm\!0.33$ & $4.65\!\pm\!0.09$ & 0.872 & 1.102 & $49.57\!\pm\!0.40$ & $14.04\!\pm\!0.18$ & 0.166 & 0.214 & $43.10\!\pm\!0.53$ & $15.58\!\pm\!0.22$ \\
Qwen2.5-7B-Inst (A-MEM) & 0.286 & 0.372 & $14.17\!\pm\!0.11$ & $24.66\!\pm\!0.49$ & 1.131 & 1.431 & $18.53\!\pm\!0.11$ & $54.16\!\pm\!1.10$ & 0.733 & 0.966 & $18.87\!\pm\!0.15$ & $92.21\!\pm\!2.45$ \\
Qwen2.5-7B-Inst (truncate) & 0.261 & 0.335 & $8.32\!\pm\!0.05$ & $6.02\!\pm\!0.17$ & 0.968 & 1.228 & $11.78\!\pm\!0.10$ & $11.63\!\pm\!0.19$ & 0.396 & 0.498 & $13.28\!\pm\!0.14$ & $22.12\!\pm\!0.59$ \\
Search-R1 & 0.454 & 0.533 & $13.05\!\pm\!0.08$ & $4.29\!\pm\!0.22$ & 0.064 & 0.080 & $24.74\!\pm\!0.17$ & $4.33\!\pm\!0.16$ & 0.009 & 0.011 & $20.91\!\pm\!0.03$ & $4.66\!\pm\!0.17$ \\
DeepResearcher & 0.532 & 0.646 & $21.79\!\pm\!0.41$ & $4.01\!\pm\!0.08$ & 0.731 & 0.902 & $51.87\!\pm\!0.38$ & $11.16\!\pm\!0.16$ & 0.071 & 0.106 & $48.31\!\pm\!0.60$ & $15.79\!\pm\!0.18$ \\
MEM1-QA (7B) & 0.708 & 0.836 & $6.39\!\pm\!0.02$ & $6.49\!\pm\!0.06$ & 1.860 & 2.301 & $8.05\!\pm\!0.06$ & $8.61\!\pm\!0.11$ & 1.883 & 2.285 & $10.31\!\pm\!0.08$ & $8.71\!\pm\!0.10$ \\
\midrule
Mixed-horizon SFT (3B) & 0.469 & 0.606 & $4.85\!\pm\!0.08$ & $4.44\!\pm\!0.18$ & 0.913 & 1.146 & $6.56\!\pm\!0.33$ & $14.43\!\pm\!0.23$ & 1.048 & 1.352 & $12.26\!\pm\!0.05$ & $18.23\!\pm\!0.13$ \\
PPO (3B) & $0.658\!\pm\!0.015$ & $0.806\!\pm\!0.020$ & $6.49\!\pm\!0.03$ & $5.53\!\pm\!0.22$ & $0.538\!\pm\!0.015$ & $0.655\!\pm\!0.015$ & $7.93\!\pm\!0.65$ & $22.32\!\pm\!0.18$ & $0.442\!\pm\!0.011$ & $0.549\!\pm\!0.024$ & $14.42\!\pm\!0.16$ & $24.60\!\pm\!0.14$ \\
\method{} (3B) & $\mathbf{0.756\!\pm\!0.020}$ & $\mathbf{0.921\!\pm\!0.024}$ & $7.29\!\pm\!0.03$ & $5.91\!\pm\!0.16$ & $\mathbf{1.913\!\pm\!0.024}$ & $\mathbf{2.510\!\pm\!0.019}$ & $8.20\!\pm\!0.11$ & $20.77\!\pm\!0.14$ & $\mathbf{2.430\!\pm\!0.029}$ & $\mathbf{2.834\!\pm\!0.019}$ & $9.66\!\pm\!0.09$ & $21.14\!\pm\!0.11$ \\
\bottomrule
\end{tabular}}
\caption{Multi-objective multi-hop retrieval on Q2, Q8, and Q16. EM and F1 are summed across questions. Peak context is in $10^2$ tokens and time is in seconds per trajectory. PPO and MemOPD report mean $\pm$ standard deviation over five seeds.}
    \vspace{-0.3cm}
\label{tab:main-results}
\end{table*}

This sampled-action-domain contract is separate from state reconstruction. Correct visibility can still mark a retained response copy as another action, causing the actor, critic, and GAE to count one decision twice. Conversely, a correct action domain cannot repair logits computed with deleted history or wrong positions. State reconstruction preserves conditioning, while the action domain preserves which tokens count as decisions.

\subsection{Teacher Guidance and Policy Optimization}
Once the states and action domain are aligned, \method{} combines local teacher guidance with task-level policy optimization. The student and frozen teacher receive the same tokens, positions, and visibility and share a tokenizer and vocabulary $\mathcal V$. For a supervised action position $j$ containing token $a_{t,k}$, let $p_{\theta,j}$ and $p_{T,j}$ be the distributions read at $g(t,k)$ from the student and teacher. We use full vocabulary reverse KL divergence,
\begin{equation}
    \mathcal L_{\mathrm{OPD}}
    =\frac{1}{\sum_j m^{\mathrm{T}}_j}
    \sum_j m^{\mathrm{T}}_j
    \sum_{v\in\mathcal V}p_{\theta,j}(v)
    \log\frac{p_{\theta,j}(v)}{p_{T,j}(v)}.
    \label{eq:teacher}
\end{equation}
The divergence itself is standard in language model distillation~\citep{gu2024minillm,ko2024distillm,agarwal2024gkd}. Its role here is to expose the complete teacher distribution only after the student memory state has been recovered.

The reference policy is separate from both the teacher and the behavior snapshot. It remains frozen at the SFT initialization, whereas $\pi_{\mathrm{old}}$ is the policy that generated the current rollout batch. Following the KL regularization commonly used in reinforcement learning from human feedback~\citep{ouyang2022instructgpt}, the reference policy constrains drift by adding a token-level KL penalty to the task reward. Let $\ell_{\mathrm{old},j}$ and $\ell_{\mathrm{ref},j}$ be the log probabilities assigned to the sampled action token by the behavior and reference policies, respectively. We define the reference KL penalty and the resulting token-level reward as
\begin{equation}
    \mathrm{KL}^{\mathrm{ref}}_j
    =
    \left[
    \ell_{\mathrm{old},j}
    -
    \ell_{\mathrm{ref},j}
    \right]_+,
    \qquad
    \widetilde r_j
    =
    r_j-\beta\,\mathrm{KL}^{\mathrm{ref}}_j.
    \label{eq:reward}
\end{equation}
Here, $\beta$ controls reference regularization. The scalar task reward is placed on the final sampled action token, while the reference penalty is defined at every sampled action position. The critic predicts values at these positions, and GAE propagates the resulting signal backward through the ordered student actions. Let $\tau_1,\ldots,\tau_N$ be the positions for which $m^{\mathrm{act}}=1$. The recursion follows these decisions rather than adjacent storage positions:
\begin{equation}
\begin{aligned}
    \delta_n &= \widetilde r_{\tau_n}
    +\gamma V_{\tau_{n+1}}-V_{\tau_n},\\
    A_{\tau_n} &= \delta_n+\gamma\lambda_{\mathrm{GAE}}A_{\tau_{n+1}}.
\end{aligned}
\label{eq:gae}
\end{equation}
Here, $\gamma$ is the discount factor and $\lambda_{\mathrm{GAE}}$ is the GAE decay parameter. The value after a terminal action is zero. Context copies and observations do not introduce additional decisions or value targets.

For each sampled action position $j$ containing token $a_{t,k}$, let $\sigma_j=\sigma_{t,k}$ and $\widetilde\sigma_j=\widetilde\sigma_{t,k}$. Because another state would mix a policy change with a change in conditioning, both policies evaluate $a_j=a_{t,k}$ at $g(t,k)$ under the aligned state $\widetilde\sigma_j=\sigma_j$. We therefore define $\varrho_j(\theta)=\pi_\theta(a_j\mid\widetilde\sigma_j)/\pi_{\mathrm{old}}(a_j\mid\widetilde\sigma_j)$ and $\bar\varrho_j=\operatorname{clip}(\varrho_j,1-\epsilon,1+\epsilon)$, where $\epsilon$ is the PPO clipping threshold. The PPO loss is
\begin{equation}
\mathcal L_{\mathrm{PPO}}
=-\frac{1}{\sum_j m^{\mathrm{act}}_j}
\sum_j m^{\mathrm{act}}_j
\min\!\left(\varrho_jA_j,\bar\varrho_jA_j\right).
\label{eq:ppo}
\end{equation}
The combined actor objective is
\begin{equation}
    \mathcal L_{\mathrm{actor}}
    =\mathcal L_{\mathrm{PPO}}
    -\alpha\mathcal H
    +\lambda\mathcal L_{\mathrm{OPD}},
    \label{eq:actor}
\end{equation}
where $\mathcal H$ is policy entropy, and $\alpha$ and $\lambda$ control entropy regularization and teacher guidance, respectively. The teacher supplies dense local preferences, but it does not determine whether a sequence of memory updates and retrieval actions succeeds. PPO retains the task reward as the supervision signal for complete interaction success and can counteract local teacher preferences that harm the final answer.

\section{Experiments}
\subsection{Experimental Setup}
\paragraph{Benchmarks.}
Following MEM1, we use a multi-objective question answering benchmark constructed from HotpotQA and Natural Questions~\citep{zhouMEM1LearningSynergize2025,yang-etal-2018-hotpotqa,kwiatkowski-etal-2019-natural}. Each example combines several questions into one query, and the agent must retrieve evidence over multiple turns from a local Wikipedia corpus before answering them all. Q2, Q8, and Q16 contain 2, 8, and 16 questions, creating progressively longer retrieval and memory update sequences. PPO and \method{} are optimized on Q2, while Q8 and Q16 test transfer to longer horizons.

\begin{table*}[t]
\centering
\scriptsize
\setlength{\tabcolsep}{4.0pt}
\renewcommand{\arraystretch}{1.05}
\begin{tabular}{lccccc}
\toprule
System & EM $\uparrow$ & F1 $\uparrow$ & Peak $\downarrow$ & Dependency $\downarrow$ & Time $\downarrow$ \\
\midrule
Qwen2.5-7B-Inst (truncate) & 0.296 & 0.386 & $6.31\!\pm\!0.05$ & $1.647\!\pm\!0.04$ & $2.241\!\pm\!0.04$ \\
Qwen2.5-7B-Inst (A-MEM) & 0.246 & 0.372 & $8.51\!\pm\!0.11$ & $0.932\!\pm\!0.03$ & $11.319\!\pm\!0.36$ \\
Qwen2.5-7B-Inst & 0.271 & 0.390 & $9.26\!\pm\!0.21$ & $1.151\!\pm\!0.05$ & $2.302\!\pm\!0.04$ \\
Qwen2.5-14B-Inst & 0.431 & \textbf{0.533} & $8.99\!\pm\!0.20$ & $2.260\!\pm\!0.09$ & $6.870\!\pm\!0.20$ \\
Search-R1 & \textbf{0.452} & 0.518 & $11.12\!\pm\!0.27$ & $1.473\!\pm\!0.05$ & $2.218\!\pm\!0.14$ \\
DeepResearcher & 0.428 & 0.512 & $13.43\!\pm\!0.36$ & $6.844\!\pm\!0.32$ & $3.780\!\pm\!0.08$ \\
MEM1-QA (SFT) & 0.302 & 0.356 & $6.57\!\pm\!0.05$ & $3.296\!\pm\!0.13$ & $4.943\!\pm\!0.20$ \\
MEM1-QA & 0.409 & 0.470 & $\mathbf{5.62\!\pm\!0.03}$ & $0.742\!\pm\!0.02$ & $3.751\!\pm\!0.07$ \\
\midrule
PPO (3B) & $0.409\!\pm\!0.020$ & $0.486\!\pm\!0.021$ & $6.00\!\pm\!0.05$ & $1.049\!\pm\!0.190$ & $2.781\!\pm\!0.053$ \\
\method{} (3B) & $0.434\!\pm\!0.022$ & $0.522\!\pm\!0.030$ & $6.13\!\pm\!0.05$ & $\mathbf{0.717\!\pm\!0.028}$ & $\mathbf{1.988\!\pm\!0.048}$ \\
\bottomrule
\end{tabular}
\caption{Answer quality and efficiency on Wiki-RAG. Peak context and dependency are in $10^2$ and $10^5$, respectively, and time is in seconds per trajectory. PPO and MemOPD report mean $\pm$ standard deviation over five seeds.}
    \vspace{-0.2cm}
\label{tab:wiki-transfer}
\end{table*}

We also evaluate single-objective Wiki-RAG~\citep{jin2025search}, where the agent repeatedly retrieves passages from a Wikipedia datastore before answering one open-domain question. This benchmark tests transfer from multi-objective retrieval to a standard single-objective setting.

\paragraph{Metrics.}
Exact match (EM) and token F1 measure answer quality. For Q2, Q8, and Q16, both metrics are summed over the questions in each example. We also report peak context, dependency, and inference time per trajectory. PPO and \method{} results are the mean and standard deviation over five independently trained seeds; efficiency uncertainty for fixed models is the standard error across test trajectories. Dependency is $\sum_s |s|(|p_s|+\lfloor |s|/2\rfloor)$ over generated segments $s$, where $p_s$ is the visible prefix preceding that segment.

\paragraph{Models and training.}
Our student is Qwen2.5 3B~\citep{qwen2.5}. We use gpt-oss-120b~\citep{openai2025gptoss120bgptoss20bmodel} to generate agent trajectories, retaining only those with correct final answers and valid interaction formats. We convert each response turn into the corresponding model call format, yielding 20,036 turn-level examples. The student receives one epoch of SFT on this data before PPO or \method{} training. Across five seeds, PPO and \method{} use the same initialization, data order, reconstructed packed calls, sampled-action mask, and evaluation protocol. \method{} additionally uses the frozen 7B teacher with $\lambda=0.02$, selected on the Q2 development evaluation. For the matched Q2 control, PPO omits teacher guidance, while the two OPD variants use the same frozen teacher and coefficient and differ only in whether teacher scoring uses persistent history or reconstructed call states. All training runs use four NVIDIA H200 GPUs.

\subsection{Multi-Objective Retrieval}
Table~\ref{tab:main-results} compares our 3B training variants with established systems on Q2, Q8, and Q16.

\method{}-3B achieves the highest EM and F1 at all horizons. Averaged over five seeds, \method{} improves F1 over PPO by 14.3\%, 283.2\%, and 416.2\% on Q2, Q8, and Q16, respectively. On Q16, it reduces peak context by 33.0\% and inference time by 14.1\%. The growing advantage shows that aligned teacher guidance is more useful at longer horizons.

\subsection{Single-Objective Wiki-RAG Transfer}
Table~\ref{tab:wiki-transfer} compares the resulting PPO and \method{} models with established systems on Wiki-RAG.

On Wiki-RAG, \method{} improves EM and F1 over PPO by 6.1\% and 7.4\%, respectively, when averaged over five seeds. It also has the lowest dependency and inference time in the table. Relative to PPO, \method{} reduces dependency by 31.6\% and inference time by 28.5\%, while peak context increases by 2.2\%.

\subsection{State Alignment Verification}
We next test the mechanism behind these gains by scoring the same recorded actions under persistent history, independent invocations, and reconstructed packing. Independent invocations reproduce the original call states and serve as the reference, allowing us to measure how each training representation changes the action probabilities. The audit contains 64 real trajectories, 199 model invocations, and 12,083 sampled action tokens.
\begin{table}[t]
\centering
\scriptsize
\setlength{\tabcolsep}{3.0pt}
\renewcommand{\arraystretch}{1.06}
\begin{tabular}{lccc}
\toprule
Representation & p99 $|\Delta\log p|\downarrow$ & Top-1 $\uparrow$ & False clip $\downarrow$ \\
\midrule
Persistent history & 1.7738 & 94.61 & 13.29 \\
Batching numerical floor & $5.31\!\times\!10^{-5}$ & 100.00 & 0.00 \\
Reconstructed packing & $3.43\!\times\!10^{-5}$ & 100.00 & 0.00 \\
\bottomrule
\end{tabular}
\caption{State reconstruction on 64 student trajectories. Errors compare the same sampled actions with independent invocations in float32. False clipping is the fraction of unchanged-policy ratios outside the PPO clipping interval.}
\label{tab:correctness}
\end{table}

\begin{table}[t]
\centering
\scriptsize
\setlength{\tabcolsep}{2.2pt}
\renewcommand{\arraystretch}{1.06}
\begin{tabular}{lccc}
\toprule
Context update & Tokens & p99 $|\Delta\log p|\downarrow$ & Top-1 $\uparrow$ \\
\midrule
Full response & 3,329 & $3.43\!\times\!10^{-5}$ & 100.00 \\
Suffix retention & 3,329 & $3.83\!\times\!10^{-5}$ & 100.00 \\
Summary replacement & 3,329 & $3.59\!\times\!10^{-5}$ & 100.00 \\
Sliding window & 3,329 & $3.72\!\times\!10^{-5}$ & 100.00 \\
Retrieval refresh & 3,329 & $3.81\!\times\!10^{-5}$ & 100.00 \\
MEM1 & 12,083 & $3.43\!\times\!10^{-5}$ & 100.00 \\
\bottomrule
\end{tabular}
\caption{RCE across five controlled context updates and native MEM1 trajectories.}
    \vspace{-0.5cm}
\label{tab:context-rules}
\end{table}

\begin{figure*}[t]
    \centering
    \includegraphics[width=\textwidth]{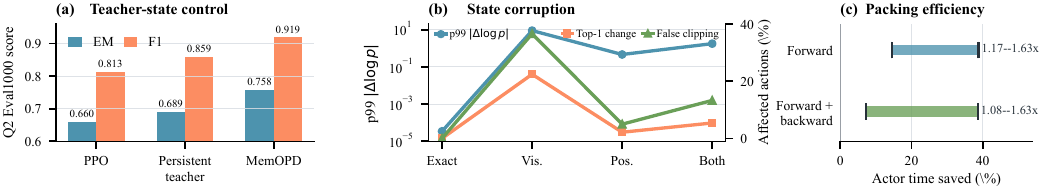}
    \caption{Experiment diagnostics. (a) Q2 results for PPO, persistent-history OPD, and \method{}. (b) Log probability error, top-1 changes, and false clipping under state corruption. (c) Actor time saved relative to independent invocations.}
    \label{fig:experiment-diagnostics}
    \vspace{-0.3cm}
\end{figure*}

Persistent history changes the top prediction at 651 action positions and falsely clips 13.29\% of the actions even before any policy update. Reconstructed packing instead matches independent invocations at the numerical floor. The mismatch also changes the supervision from the actual 7B teacher. Before the first memory update, exact and persistent states give identical teacher distributions. Across later invocations, however, top-1 agreement falls to 81.20\%, the p99 sampled log probability difference reaches 19.854, and the mean KL divergence reaches 0.793, changing the teacher prediction at 444 action positions. Having established this effect on MEM1 trajectories, we next test whether the same reconstruction interface supports other forms of context update.


Table~\ref{tab:context-rules} tests whether the interface depends on one memory topology. In every case, the compiler receives the realized token context of each invocation rather than symbolic memory roles. It therefore preserves RCE across several context updates, while MEM1 remains the complete downstream training instance.

\subsection{Ablation Studies}

\paragraph{Matched teacher-state control.}
We compare PPO, OPD with persistent-history teacher scoring, and \method{} under the same Q2 protocol. PPO uses reconstructed student-side calls and the sampled-action mask without teacher guidance. The persistent-history variant adds the frozen teacher but scores sampled actions under a flattened interaction history, whereas \method{} scores them under reconstructed call states. All other training and evaluation settings are fixed. As shown in Figure~\ref{fig:experiment-diagnostics}a, persistent-history teacher scoring improves F1 and EM over PPO by 5.6\% and 4.4\%, respectively, showing that dense teacher guidance remains useful despite the state mismatch. Reconstructing the teacher state further improves F1 and EM by 7.0\% and 10.0\%, respectively. Overall, \method{} improves F1 and EM over PPO by 13.0\% and 14.8\% in this matched control.

\paragraph{Reconstruction requirements.}
Figure~\ref{fig:experiment-diagnostics}b independently corrupts visibility and positions while keeping the recorded actions fixed. Visibility is the larger source of error, but incorrect positions alone still change 260 top predictions and falsely clip 4.97\% of the actions. State reconstruction is also distinct from identifying which tokens are decisions. The exact action domain selects 12,083 sampled action tokens, whereas trajectory and response masks add 66,830 and 74,357 spurious positions.

\paragraph{Packing Efficiency. }
Figure~\ref{fig:experiment-diagnostics}c compares reconstructed packing with correct independent invocations across nine configurations. Packing achieves up to a $1.63\times$ actor speedup while preserving RCE. This speedup comes from shared computation without changing the optimized states.

\section{Discussion}
\method{} separates action provenance from state validity. A student rollout guarantees that the learner generated each action, but compact memory may delete, replace, or reencode its conditioning context before training. The resulting batch can therefore contain genuine student actions evaluated under states that never occurred during interaction. Memory state alignment enforces this requirement, and RCE audits the reconstructed logits against independent invocations.

Teacher guidance and PPO serve different roles in the objective. The teacher provides dense guidance at sampled action tokens, whereas PPO evaluates whether the complete interaction solves the task. The matched control makes this distinction measurable: persistent-history teacher scoring improves F1 over PPO by 5.6\%, showing that teacher guidance remains useful before state alignment, while reconstructing the teacher state adds a further 7.0\% improvement.

Beyond the optimization objective, state correctness also determines how packing can be used. \method{} restores each invocation before sharing stable prefixes across otherwise independent computations. RCE verifies that this sharing preserves the action logits, so the measured speedup does not depend on an approximate rollout state.

The same reconstruction interface is not tied to semantic roles defined by MEM1. Because the compiler consumes realized token contexts rather than a fixed memory topology, it preserves RCE under full response retention, suffix retention, summary replacement, sliding windows, and retrieval refresh. These results show that memory state alignment depends on the actual computation presented to the model rather than the particular form of memory rewriting, allowing \method{} to support diverse context update mechanisms through a unified training interface.

\section{Conclusion}
Compact memory can separate sampled actions from the states later used to train them. \method{} resolves this mismatch by reconstructing every invocation, distinguishing sampled actions from later context copies, and combining state-aligned teacher guidance with task-level PPO. RCE verifies the packed computation, while the matched control shows that teacher guidance already improves F1 over PPO by 5.6\% and state alignment adds a further 7.0\%. Across five seeds, \method{} improves F1 over PPO by up to 416.2\%, while packing yields up to a $1.63\times$ speedup in actor computation during training.
\bibliography{references_state_mismatch}

\end{document}